\documentclass{article}

\usepackage[preprint]{neurips_2025}
\usepackage{subcaption}
\usepackage{hyperref}       

\usepackage[utf8]{inputenc} 
\usepackage[T1]{fontenc}    
\usepackage{hyperref}       
\usepackage{url}            
\usepackage{booktabs}       
\usepackage{amsfonts}       
\usepackage{nicefrac}       
\usepackage{microtype}      
\usepackage{xcolor}    
\usepackage{amsmath}
\usepackage{graphicx}

\usepackage{amsthm}
\usepackage{amssymb}
\usepackage[capitalize,noabbrev]{cleveref}

\theoremstyle{plain}
\newtheorem{theorem}{Theorem}[section]

\newtheorem{lemma}[theorem]{Lemma}
\newtheorem{corollary}[theorem]{Corollary}

\theoremstyle{definition}

\theoremstyle{remark}

\crefname{theorem}{Theorem}{Theorems}
\crefname{proposition}{Proposition}{Propositions}
\crefname{lemma}{Lemma}{Lemmas}
\crefname{corollary}{Corollary}{Corollaries}
\crefname{definition}{Definition}{Definitions}
\crefname{assumption}{Assumption}{Assumptions}
\crefname{remark}{Remark}{Remarks}
\title{The Effect of Depth on the Expressivity of Deep Linear State-Space Models}

\author{%
  Zeyu Bao \\
  Department of Mathematics \\
  National University of Singapore \\
  \texttt{zeyu@u.nus.edu} \\
  \And
  Penghao Yu \\
  Department of Mathematics \\
  National University of Singapore \\
  \texttt{e1353366@u.nus.edu} \\
  \And
  Haotian Jiang \\
  Department of Mathematics \\
  Institute for Functional Intelligent Materials \\
  National University of Singapore \\
  \texttt{haotian@nus.edu.sg} \\
  \And
  Qianxiao Li \\
  Department of Mathematics \\
  Institute for Functional Intelligent Materials \\
  National University of Singapore \\
  \texttt{qianxiao@nus.edu.sg} \\
}

\begin{document}

\maketitle

\begin{abstract}


Deep state-space models (SSMs) have gained increasing popularity in sequence modelling. While there are numerous theoretical investigations of shallow SSMs, how the depth of the SSM affects its expressiveness remains a crucial problem. In this paper, we systematically investigate the role of depth and width in deep linear SSMs, aiming to characterize how they influence the expressive capacity of the architecture. First, we rigorously prove that in the absence of parameter constraints, increasing depth and increasing width are generally equivalent, provided that the parameter count remains within the same order of magnitude. However, under the assumption that the parameter norms are constrained, the effects of depth and width differ significantly. We show that a shallow linear SSM with large parameter norms can be represented by a deep linear SSM with smaller norms using a constructive method. In particular, this demonstrates that deep SSMs are more capable of representing targets with large norms than shallow SSMs under norm constraints. Finally, we derive upper bounds on the minimal depth required for a deep linear SSM to represent a given shallow linear SSM under constrained parameter norms. We also validate our theoretical results with numerical experiments.











\end{abstract}

\section{Introduction}
Recent advances in state space model(SSM) have been successful in learning long sequence relationships via mitigating the computational inefficiency of explicitly modeling token interactions\citep{gu2023mamba,gu2021efficiently,gu2020hippo,gu2021combining,smith2022simplified}. It achieves significantly better performance compared with attention-based transformers in the long range arena (LRA) dataset \citep{tay2020long}. 
The linear time-invariant structure of SSM allows for an asymptotic computational complexity of only $O(T\log T)$, which is significantly better than the $O(T^{2})$ complexity of traditional full-attention approaches exhibiting significant computational demands \citep{vaswani2017attention}. Moreover, SSMs have proven effective in multiple domains dealing with continuous signal data, including audio and vision tasks \citep{li2024videomamba,goel2022s,nguyen2022s4nd}.

Despite its success across various fields of practical application, many theoretical questions remain unanswered. One of the biggest problems is the expressivity of deep state-space models. Although modern SSMs \citep{smith2022simplified,gu2023mamba} comprise dozens of layers and contain millions of parameters, it is still unclear what the real difference is between shallow SSMs and deep SSMs, and the reason why we need depth in state-space models. Another notable problem is what kind of sequence to sequence relationship can be handled better with a deep network than with a shallow network.





A number of recent works have employed dynamical systems approaches to study token stability in deep SSMs\citep{vo2024demystifying}. \cite{geshkovski2023emergence} uses a similar dynamics-based formulation to investigate feature clustering behaviors in transformers. Other studies focus on the training dynamics of deep linear networks from an optimization perspective \citep{menon2024geometry}. However, relatively little work has been devoted to understanding how depth affects the expressivity of deep SSMs. In this paper, we adopt a simple formulation that allows us to characterize important differences between multi-layer SSM and one-layer SSM. Our main contributions are presented as follows:
\begin{itemize}
    \item In \cref{theorem 4.1}, we prove that, in the absence of norm constraints, depth and width are equivalent in approximation in the sense that under a fixed parameter budget, models of arbitrary depth achieve the same expressive power.
    \item In \cref{theorem 4.2}, we demonstrate that a shallow SSM with large-norm weights can be exactly represented by a deep SSM with smaller-norm weights, under explicit norm constraints on the model parameters.
    \item We establish upper bounds on the minimal depth required for a deep linear SSM to represent a given shallow linear SSM in the regime of constrained parameter norms in \cref{theorem 4.3} and validate our theoretical results with numerical experiments.
\end{itemize}





\section{Related works}

\paragraph{Expressivity of State-Space Models} State-space models originate from the HIPPO matrix, which is optimal in the online function approximation sense \citep{smith2022simplified,gu2020hippo}. \citet{wang2023state} also provides theoretical guarantees for the approximation of continuous sequence-to-sequence mappings using SSMs with layer-wise nonlinear activations. Furthermore, \cite{muca2024theoretical} employed tools from Rough Path Theory to show that deep diagonal SSMs possess less expressive power than their non-diagonal counterparts. 
However, these works primarily focus on function approximation of deep SSMs, whereas the present work aims to present the simplest setting where one can study how depth influences the expressivity in deep SSMs.


\paragraph{Dynamics in deep State-Space Models} Several works focus on dynamics in deep SSMs. \cite{vo2024demystifying} investigates the divergence behavior of tokens in a pre-trained Mamba model by characterizing continuous-time systems. \cite{smekal2024interplay} shows how the memory of a deep linear SSM varies with the depth and width, and how the learning dynamics of deep linear models vary with memory expressivity. Our work focuses on a different setting and complements existing analyses by providing a theoretical understanding of depth in deep linear SSMs.



\paragraph{Deep linear networks} Simple linear architectures often serve as effective tools for gaining theoretical insights into the behavior of deep neural networks. \cite{arora2018convergence} proves that the convergence of gradient descent achieves a linear rate for training a deep linear neural network over whitened data under certain conditions. \cite{bah2022learning} also shows that optimizing a deep linear network is equivalent to Riemannian gradient flow on a manifold of low-rank matrices, with a suitable Riemannian metric. \citet{gruber2024role} study the implicit bias arising from weight initialization in deep linear networks, offering the insight that weight norms are central to understanding the behavior of deep models. Following prior theoretical analysis of deep linear networks, our work focuses on understanding the expressivity of deep recurrent models, specifically deep state-space models.






\section{Problem Formulation}

One particularly notable feature of currently popular sequential models such as S4\citep{smith2022simplified} and Mamba\citep{gu2023mamba} is the stacking of multiple layers, which leverages scaling laws to increase the number of parameters and achieve stronger performance. In this section, we aim to provide a theoretical perspective on the architecture of deep state-space models, to characterize more precisely how depth contributes to their success.

\subsection{Deep Linear State-Space Models}



For a general $l$-layer deep state-space model, we express its recurrent form as follows, where each layer's input comes from the previous layer's output:
\begin{equation}
\label{linear deep ssm}
\begin{aligned}
y(t) &= C^{T}\sigma(h_{l}(t))\\
h_{l}(t+1) &= A_{l}h_{l}(t)+B_{l}\sigma(h_{l-1}(t+1))\\
&\vdots\\
h_{2}(t+1) &= A_{2}h_{2}(t)+B_{2}\sigma(h_{1}(t+1))\\
h_{1}(t+1) &= A_{1}h_{1}(t)+B_{1}x(t+1)\\
\end{aligned}
\end{equation}
where $h_{l}(t)\in\mathbb{R}^{m}$ is the hidden state of $l$-th layer at time step $t$. $\sigma:\mathbb{R}^{m\times 1}\to\mathbb{R}^{m\times1}$ is activation function on hidden state $h_{l}(t)$. $A_{1},...,A_{l}, B_{2},...,B_{l}\in\mathbb{C}^{m\times m}$ and $C,B_{1}\in\mathbb{C}^{m\times1}$ are hidden matrices. Also, $A_{1},...,A_{l}$ are the state-space matrices, $B_{1},...,B_{l}$ are the input matrices and $C$ is the output matrix. For each scalar input $x(t)\in\mathbb{R}$, we obtain an output $y(t)\in\mathbb{R}$ by passing it through this $l$-layer model.


Now, we set $\sigma$ to be the identity mapping and consider a linear setting. We observe that both shallow and deep state-space models share a convolutional structure, differing only in their kernels when the nonlinearity in the function $\sigma$ is removed, which serves as a natural and tractable starting point for investigating the role of depth in SSMs.

The convolutional kernel allows us to clearly understand how the input is mapped to the corresponding output through the sequential model, i.e. $y(t)=(\rho \ast x) (t)$. We characterize the convolutional kernel of the deep state-space model as follows:

\begin{lemma} \label{lemma: convolution kernel}
     For $l$ layer deep linear SSM defined in \cref{linear deep ssm}, the convolutional kernel $\rho(t)$ admits the following representation
\begin{equation}
\rho(t)=\sum_{\substack{i_{1}+i_{2}+...+i_{l}=t\\i_{1},...,i_{l}\in\mathbb{N}}}C^{T}\prod_{j=1}^{l}(A_{l-j+1}^{i_{l-j+1}}B_{l-j+1})
\end{equation}
\end{lemma}

A proof of \cref{lemma: convolution kernel} can be found in \cref{ProofOfStructure}. Our approach to proving this result is checking the basic case and then performing induction on both timesteps $t$ and layers $l$. 

Notably, it is the simplified linear architecture of our network that enables the convolutional kernel to be expressed in a clear and tractable form, as the sum over layer indices of successive powers of the hidden matrices. This lemma reveals how, in deep SSMs, each input $x(t)$ is convolved through such a kernel to produce the output $y(t)$. This insight lays the groundwork for our subsequent comparison of the representational capacities of deep SSMs and shallow SSMs.

\subsection{Norm-Constrained Hypothesis Space for Deep Linear SSMs}
\label{space definition}
To provide a mathematical description of the function space of deep state-space models, leveraging \cref{lemma: convolution kernel}, we define the following hypothesis spaces for deep linear state-space models under norm constraints.
\begin{equation}
\label{4}
\begin{aligned}
    \mathcal{H}_{c,l}^{m}=&\{\rho(t):y(t)=(\rho \ast x) (t),A_{1},...,A_{l}\in\mathbb{C}^{m\times m}\thinspace\mbox{diagonal},B_{2},...,B_{l},\in\mathbb{C}^{m\times m},\\ &C,B_{1}\in\mathbb{C}^{m\times1},\max_{i=1,...,l}r(A_{i})<1,||C||_{\infty}\le c, ||B_{1}||_{\infty}\le c,\max_{2\le k\le l}\max_{1\le i,j\le m}|(B_{k})_{ij}|\le c\}
\end{aligned}
\end{equation}
where $r(\cdot)$ refers to spectral radius of each matrix, $||\cdot||$ means the infinity norm of a vector, $m$ is the width of network, $c$ is norm constraint for parameters and $l$ is the number of layers.

Here, we assume that each $A_i$ is diagonal, a structure commonly adopted in real-world models \citep{saon2023diagonal}. We further impose that the spectral radius of each $A_i$ is strictly less than 1 to ensure system stability. Since the implementation in \cite{smith2022simplified} is one-dimensional, we assume both the input $x(t)$ and output $y(t)$ are scalar-valued. It is a natural thing to consider norm constraints because of optimization. State-space models may suffer from large parameter norms, which may lead to training instability and negatively impact model performance \citep{pascanu2013difficulty}. Thus, the norm constraint $c$ remains an important factor in distinguishing the expressivity of deep state-space models, for the simple reason that the convolutional kernel is directly determined by the parameters $B_{i}$ and $C$. Thus, altering the norm can have a significant impact on the model’s expressivity.

\section{Main Results}
In this section, we present our main results on the expressivity of deep state-space models within the framework defined in \cref{space definition}.

\subsection{Equivalence of Depth and Width without Norm Constraints}
\label{subsection 4.1}
First, we focus on the hypothesis space of deep linear SSMs without norm constraints. Our goal is to characterize the fundamental differences between shallow and deep SSMs, which leads us to the following questions: given a one-layer SSM with a certain width, 
how wide must a $l$-layer SSM be in order to represent it? In previous work, \citet{smekal2024interplay} provided an example demonstrating how a four-layer linear SSM can be converted into a single-layer one. Here, we give a complete characterization.



\begin{theorem} 
\label{theorem 4.1}
Let $m,l\geq 1$ and $c_{1}>0$. Recall that $\mathcal{H}_{\infty,l}^{m}$ is a norm-unconstrained hypothesis space of linear SSMs with $l$ layers and $m$ width. Then, we have
\begin{equation}
\begin{aligned}
    &\mathcal{H}_{\infty,1}^{l(m-1)+1}\subseteq\mathcal{H}_{\infty,l}^{m}\subseteq\mathcal{H}_{\infty,1}^{lm}\\
    &\mathcal{H}_{\infty,1}^{l(m-1)+2}\not\subseteq\mathcal{H}_{\infty,l}^{m}
\end{aligned}
\end{equation}
\end{theorem}



A detailed proof of \cref{theorem 4.1} based on explicit construction can be found in \cref{ProofMain}. 

On the one hand, \cref{theorem 4.1} shows that given an $l$-layer linear SSM with width $m$, we can always construct a one-layer linear SSM with width $ml$ to represent it.
On the other hand, the maximal width of one-layer SSM that can be represented by an $l$-layer SSM with width $m$ is $l(m-1)+1$. This implies that there exists a convolutional kernel of one-layer linear SSM with width $l(m-1)+2$ that cannot be represented by the kernel of an $l$-layer SSM with width $m$. We have constructed such a kernel to illustrate the optimality of this width bound, as detailed in \cref{ProofMain}. This result illustrates that $l$-layer SSM of width $m$ has expressivity equivalent to one-layer SSM of width $O(lm)$ under the same parameter count, highlighting that width can be traded for depth without loss of expressive power.
\par In fact, a more general result holds when the assumption that each hidden matrix $A_i$ is diagonal is relaxed to the case where $A_i$ is diagonalizable, which is a dense and open set in the matrix space $\mathbb{C}^{m\times m}$. See the appendix \cref{ProofMain} for details.


\subsection{Non-equivalence of Depth and Width with Norm Constraints}
\label{section 4.2}
In the absence of norm constraints, increasing width and depth are equivalent under the same parameter count (\cref{theorem 4.1}). However, under norm constraints, the effects of depth and width on the expressivity of deep linear SSMs differ significantly. Now, we focus on the hypothesis space defined in \cref{space definition} to investigate the impact of norm constraints on the expressivity of deep linear state-space models. The following theorem demonstrates how a shallow SSM with a large parameter norm can be equivalently represented by a deeper SSM composed of layers with smaller parameter norms:

\begin{theorem}
\label{theorem 4.2} Suppose $m,l\ge1$ and let $c_{1}>0$ be the norm constraint for the one-layer linear SSM. Then, we have the following upper bound for the norm constraint of $l$-layer linear SSM defined in \cref{linear deep ssm}:
    \begin{equation}
    \sup_{\rho\in\mathcal{H}_{c_{1},1}^{l(m-1)+1}}\inf_{c_{2}>0}\{c_{2},\: \rho\in\mathcal{H}_{c_{2},l}^{m}\}\le2c_{1}^{\frac{2}{l+1}}
\end{equation}
\end{theorem}

A detailed proof of \cref{theorem 4.2} can be found in \cref{ProofMain}. Notably, under the same order of magnitude of parameter count, a given one-layer linear SSM with a large norm constraint $c_{1}$ can be equivalently represented by an $l$-layer SSM with width $m$ and a smaller norm constraint bounded by $2c_{1}^{\frac{2}{l+1}}$. Hence, as the number of layers increases, the corresponding norm of weights decreases very quickly, indicating that depth plays an important role in reducing the norm from the approximation perspective.


Here we give an example showing the construction of converting a known $1$ layer linear SSM with width $2m-1 = 7$ into a $2$ layer linear SSM with width $m=4$. Given distinct non-zero complex numbers $|\alpha_1| \le \dots \le |\alpha_7|$, we suppose that the $1$ layer SSM is defined by $B = C = (z_1, \dots, z_7)^T$ and $A = Diag\{\alpha_1, \dots, \alpha_7\}$. Let $c_0 = \max_{1 \le i \le 7} |z_i|$ and the $\rho$ defined by the above SSM satisfies $\rho \in \mathcal{H}_{c_{0},1}^{7}$, with $\rho(t) = \sum_{i=1}^7 z_i^2 \alpha_i^t$. 
\par Then, we consider constructing a $2$ layer SSM with the same $\rho$. Let $Z_0 = 2 c_0^{\frac{2}{3}}$. The following constructed $A_1, A_2, B_1, B_2, C$ defines the corresponding $2$ layer SSM of width $4$. \begin{itemize}
    \item $A_1 = Diag \{\alpha_1, \alpha_2, \alpha_3, \alpha_4\}$. 
    \item $A_2 = Diag\{\alpha_5, \alpha_6, \alpha_{7}, 0\}$. 
    \item $B_1 = C = (Z_0, Z_0, Z_0, Z_0)^T$.
    \item $B_2 = \begin{pmatrix}
 \frac{(\alpha_5 - \alpha_1)z_5^2}{\alpha_5 Z_0^2} & 0 &  0 & 0\\
 0 & \frac{(\alpha_6 - \alpha_2)z_6^2}{\alpha_6 Z_0^2} &   0& 0\\
 0 & 0 &  \frac{(\alpha_7 - \alpha_3)z_7^2}{\alpha_7 Z_0^2} & 0\\
\frac{(z_{1}^2) + (z_{5}^2)\frac{\alpha_1}{\alpha_5}}{Z_0^2}  & \frac{(z_{2}^2) + (z_{6}^2)\frac{\alpha_2}{\alpha_6}}{Z_0^2} &  \frac{(z_3^2) + (z_7^2)\frac{\alpha_{3}}{\alpha_7}}{Z_0^2} & \frac{z_{4}^2}{Z_0^2}
\end{pmatrix}$
\end{itemize} 
Then by calculating the convolutional kernel $\hat{\rho}$ of the above $2$-layer SSM, we have
\begin{align}
    \hat{\rho}(t) &= z_4^2 \alpha_4^t + \sum_{i = 1}^3 ([{(z_{i}^2) + (z_{i+4}^2)\frac{\alpha_i}{\alpha_{i+4}}}] \alpha_{i}^t +  \frac{(\alpha_{i+4} - \alpha_i)z_{i+4}^2}{\alpha_{i+4}} \sum_{s = 0}^t \alpha_i^{s} \alpha_{i+4}^{t-s})\\
    &= z_4^2 \alpha_4^t + \sum_{i = 1}^3 ([{(z_{i}^2) + (z_{i+4}^2)\frac{\alpha_i}{\alpha_{i+4}}} ]\alpha_{i}^t +  \frac{(\alpha_{i+4} - \alpha_i)z_{i+4}^2}{\alpha_{i+4}} \frac{\alpha_{i+4}^{t+1} - \alpha_i^{t+1}}{\alpha_{i+4} - \alpha_i})\\
    &= z_4^2 \alpha_4^t + \sum_{i = 1}^3 ([{(z_{i}^2) + (z_{i+4}^2)\frac{\alpha_i}{\alpha_{i+4}}} - (z_{i+4}^2)\frac{\alpha_i}{\alpha_{i+4}}]\alpha_{i}^t +  \frac{z_{i+4}^2 \alpha_{i+4}^{t+1}}{\alpha_{i+4}})\\
    &= z_4^2 \alpha_4^t + \sum_{i = 1}^3 ({z_{i}^2}\alpha_{i}^t + {z_{i+4}^2 \alpha_{i+4}^{t}})\\
    &= \rho(t)
\end{align}
Then the above is an explicit construction of converting a $1$ layer linear SSM with width $7$ into a $2$ layer linear SSM with width $4$ under the same parameter count with norm constraints. Actually, in the case of converting into $l$ layer SSM, the positions of non-zero elements of $B_2, \dots, B_l$ are the same as those $B_2$ above. 
\par From $|\alpha_i| \le |\alpha_{i+4}|$, all the terms on the $m$-th column of $B_2$ are bounded by $\frac{\max \{|z_i^2| + |z_{i+4}^2|\}}{Z_0^2} \le Z_0 = 2 c_0^{\frac{2}{3}}$, matching \textbf{Theorem \ref{theorem 4.2}}. The core idea is as follows: suppose that the $1$-layer SSM have weight norm $c_0$, the $l$ layer SSM decomposes the weight into a product of $l+1$ new weight matrices, each can take a norm of order $O(c_0^{\frac{2}{l+1}})$. A similar argument holds for the case where $c_0 < 1$, in which the norms of the weights of the deep SSM can be increased.  


\subsection{Minimal depth for representing shallow networks}

Following \cref{section 4.2}, we have highlighted the important role of depth in enabling norm reduction in deep linear state-space models. In this section, we address a more refined question: suppose we have a SSM with weight norms bounded by $c_1$, which is possibly a very large value, how deep must a norm-constrained deep linear SSM with norm bound $c_2$ be to achieve the same expressive capacity? In fact, very large parameter norms may arise when using SSMs to learn nonsmooth or highly oscillatory memory function\citep{pascanu2013difficulty} while $c_2$ could be predetermined based on desired stability constraints imposed by the optimizer. \cref{theorem 4.2} already indicates that substantial norm reduction is possible through increased depth and we now formalize this by deriving the minimal depth required for a deep linear SSM with norm bound $c_2$ to represent a given one-layer linear SSM constrained by $c_1$.

\begin{theorem}
\label{theorem 4.3}
Suppose $\rho \in \mathcal{H}_{c_1,1}^{K+1}$ for some $K \geq 2$, and let $c_1 > 1$ and $c_2 > 2$ denote the norm constraints for the one-layer and deep linear SSM, respectively. Then, we have the following upper bound for the depth of a deep linear SSM as defined in \cref{linear deep ssm}:
\begin{equation}
\min\{l: \:  \rho\in\mathcal{H}_{c_{2},l}^{\lceil \frac{K}{l} \rceil +1}\} \le \lceil\frac{2\ln(c_{1})}{\ln{(\frac{c_{2}}{2})}}-1\rceil
\end{equation}
\end{theorem}
A detailed proof of \cref{theorem 4.3} can be found in \cref{ProofMain}.
The technique employed to determine the minimal required depth closely follows the constructive approach developed in \cref{theorem 4.2}. 

To maintain a constant parameter count, we set the width of an $l$-layer linear SSM to be $\lceil \frac{K}{l} \rceil + 1$. If the total number of parameters is allowed to increase, the problem becomes trivial, as parameter norms can be reduced simply by allocating more parameters. However, our results in \cref{theorem 4.3} indicate that increasing the depth is an efficient strategy even under a fixed parameter budget. This suggests that model performance may be improved by increasing depth if the parameter norm is large. We will show later in our experiments \cref{experiment}.


\subsection{Beyond Diagonal Case}
Recall that \cref{theorem 4.1} extends to the case where the hidden matrices are diagonalizable. However, extending \cref{theorem 4.2} to diagonalizable matrices remains challenging due to the condition number induced by diagonalization. To obtain tractable norm bounds, we instead assume that each hidden matrix $A_i$ is normal, i.e., unitarily diagonalizable with a condition number $1$, which resolves this issue. 
This assumption also aligns with the HiPPO initialization commonly used in practice \citep{gu2020hippo}. Now we define the following hypothesis space:  
\begin{equation}
\begin{aligned}
  \mathcal{G}_{c,l}^{m}=&\{\rho(t):y(t)=(\rho\ast x)(t),A_{1},...,A_{l}\in\mathbb{C}^{m\times m}\thinspace\mbox{normal},B_{2},...,B_{l},\in\mathbb{C}^{m\times m},\\ &C,B_{1}\in\mathbb{C}^{m\times1},\max_{i=1,...,l}r(A_{i})<1,||C||_{\infty}\le c, ||B_{1}||_{\infty}\le c,\max_{2\le k\le l}\max_{1\le i,j\le m}|(B_{k})_{ij}|\le c\}
\end{aligned}
\end{equation}

The following result generalizes \cref{theorem 4.2} to the normal case.
\begin{corollary}
Suppose $m,l\ge1$ and let $c_{1}>0$ be the norm constraint for the one-layer linear SSM. Then, we have the following upper bound for the norm constraint of $l$-layer linear SSM, where the hidden matrix $A_{i}$ is normal.
\label{Hermite property}
    \begin{equation}
    \max_{\rho\in\mathcal{G}_{c_{1},1}^{l(m-1)+1}}\min_{c_{2}>0}\{c_{2},\: \rho\in\mathcal{G}_{c_{2},l}^{m}\}\le2((l(m-1)+1)c_{1}^{2})^{\frac{1}{l+1}}
\end{equation}
\end{corollary} 

A detailed proof of \cref{Hermite property}, as a generalization of \cref{theorem 4.2}, can be found in \cref{ProofMain}. It is important to note that, unlike the upper bound in \cref{theorem 4.2}, this norm upper bound explicitly depends on both the number of layers $l$ and width $m$, which reveals the fundamental difference between deep and shallow networks in their expressive ability in a larger space.
Precisely, unlike the unchanged bound in Theorem \ref{theorem 4.2}, the bound for this larger space is increasing at the rate of $m^{\frac{1}{l+1}}$, which can also be bounded by a constant if $l = O(\log(m))$. 

\section{Experiments}
\label{experiment}
In this section, we validate our theoretical results through numerical experiments. Before presenting our results, we provide \cref{expansion} that facilitates the computation of the output coefficients of a deep linear SSM as described in \cref{linear deep ssm}, i.e., its direct expansion into the equivalent one-layer SSM form.
\begin{lemma}
\label{expansion}
    Suppose in \cref{linear deep ssm}, $A_{i}=diag(\lambda_{i1},...,\lambda_{im})$ where each eigenvalue is distinct. We denote the $(i,j)$ element of matrix $B_{k}$ by $b_{ij}^{(k)}=(B_{k})_{ij}$, $C^{T}=(c_{1},\cdots,c_{m})^{T}$ and $B_{1}=(b_{1},\cdots,b_{m})^{T}$. Then, by fixing $t_{l-\tilde{l}+1} = \tilde{m}$, the coefficient of the term corresponding to the $\tilde{m}$-th eigenvalue of the $A_{\tilde{l}}$ matrix, i.e., $\lambda_{\tilde{l},\tilde{m}}^{t}$ in the output expansion is
\begin{equation}
    \xi_{\tilde{l},\tilde{m}}=\sum_{t_{1}=1}^{m}...\sum_{t_{l-\tilde{l}}=1}^{m}\sum_{t_{l-\tilde{l}+2}=1}^{m}...\sum_{t_{l}=1}^{m}\frac{c_{t_{1}}b_{t_{1}t_{2}}^{(l)}b_{t_{2}t_{3}}^{(l-1)}\cdots b_{t_{l-1}t_{l}}^{(2)}b_{t_{l}}}{\prod_{\alpha=1,\alpha\neq \tilde{l}}^{l}(1-\frac{\lambda_{\alpha,t_{l-\alpha+1}}}{\lambda_{\tilde{l},t_{l-\tilde{l}+1}}})}
\end{equation}
\end{lemma}

A detailed proof can be found in \cref{ProofMain}.

\begin{figure}[!ht]
  \centering
  \includegraphics[width=0.5\textwidth]{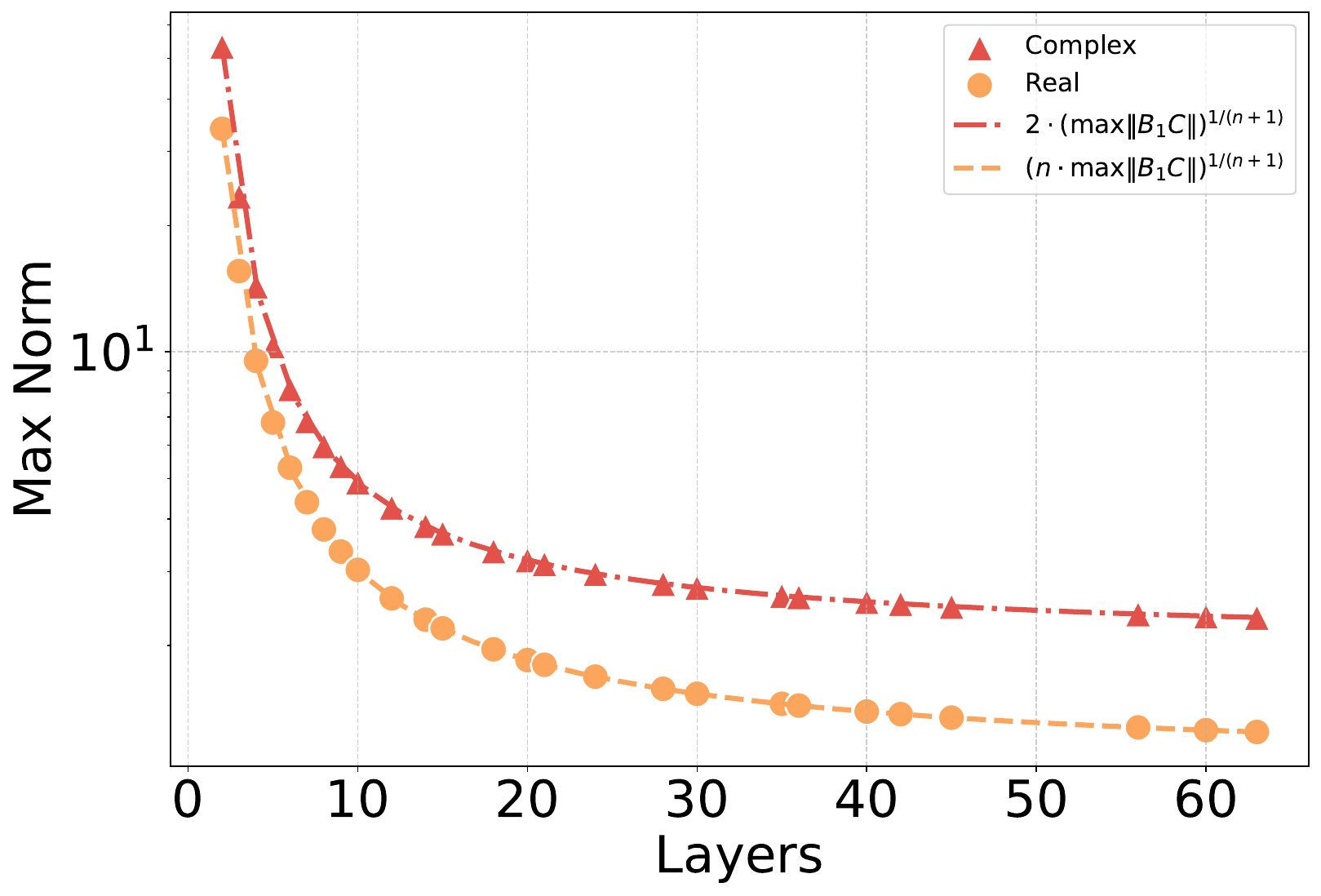}
  \caption{We use a one-layer linear network to learn a deep linear network in a teacher-student setting, examining the relationship between the number of layers and the maximum norm of the corresponding parameters in the learned model.}
  \label{fig: teacher-student}
\end{figure}

\paragraph{Numerical verification of \cref{theorem 4.2}}
We first conduct numerical experiments to verify our main theorems. Specifically, given a one-layer linear SSM with width $l(m-1)+1$, we apply the construction from \cref{theorem 4.2} to reconstruct an equivalent model with $l$ layers and width $m$. We adopt a teacher–student setup, where a wide and shallow network is learned using a deep linear network as defined in \cref{linear deep ssm}. The experiments are performed for both real-valued and complex-valued parameters. In the plot \cref{fig: teacher-student}, the dots represent the maximum observed norms, while the lines depict the theoretical bounds. Overall, the experimental results verify the construction for \cref{theorem 4.2} is correct. In particular, we observe that the maximum norms decrease at the rate predicted by \cref{theorem 4.2} as the number of layers increases. 

\begin{figure}[!ht]
    \centering
    \begin{subfigure}[b]{0.45\linewidth}
        \centering
        \includegraphics[width=1\linewidth]{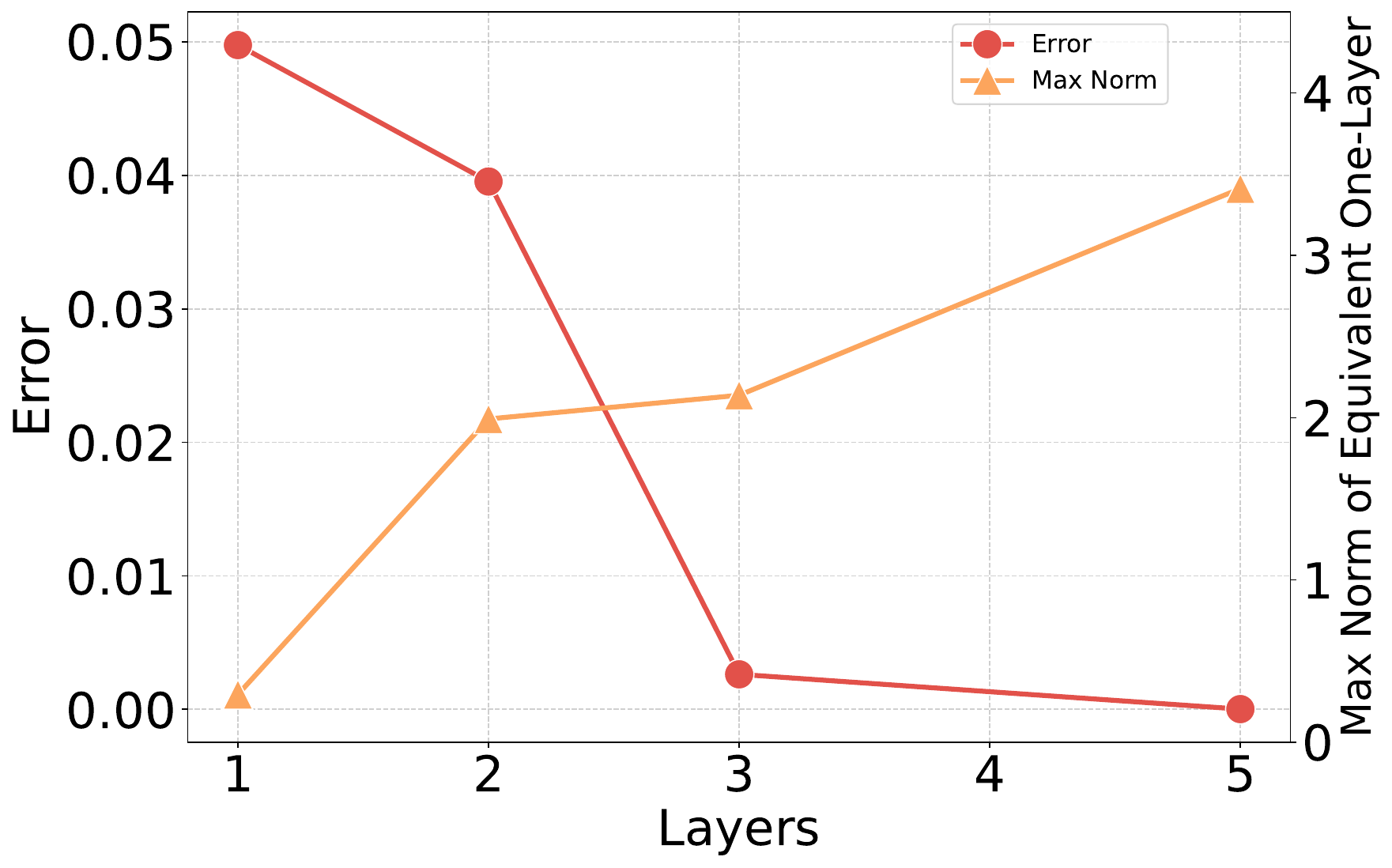}
    \end{subfigure}
    \begin{subfigure}[b]{0.42\linewidth}
        \centering
        \includegraphics[width=1\linewidth]{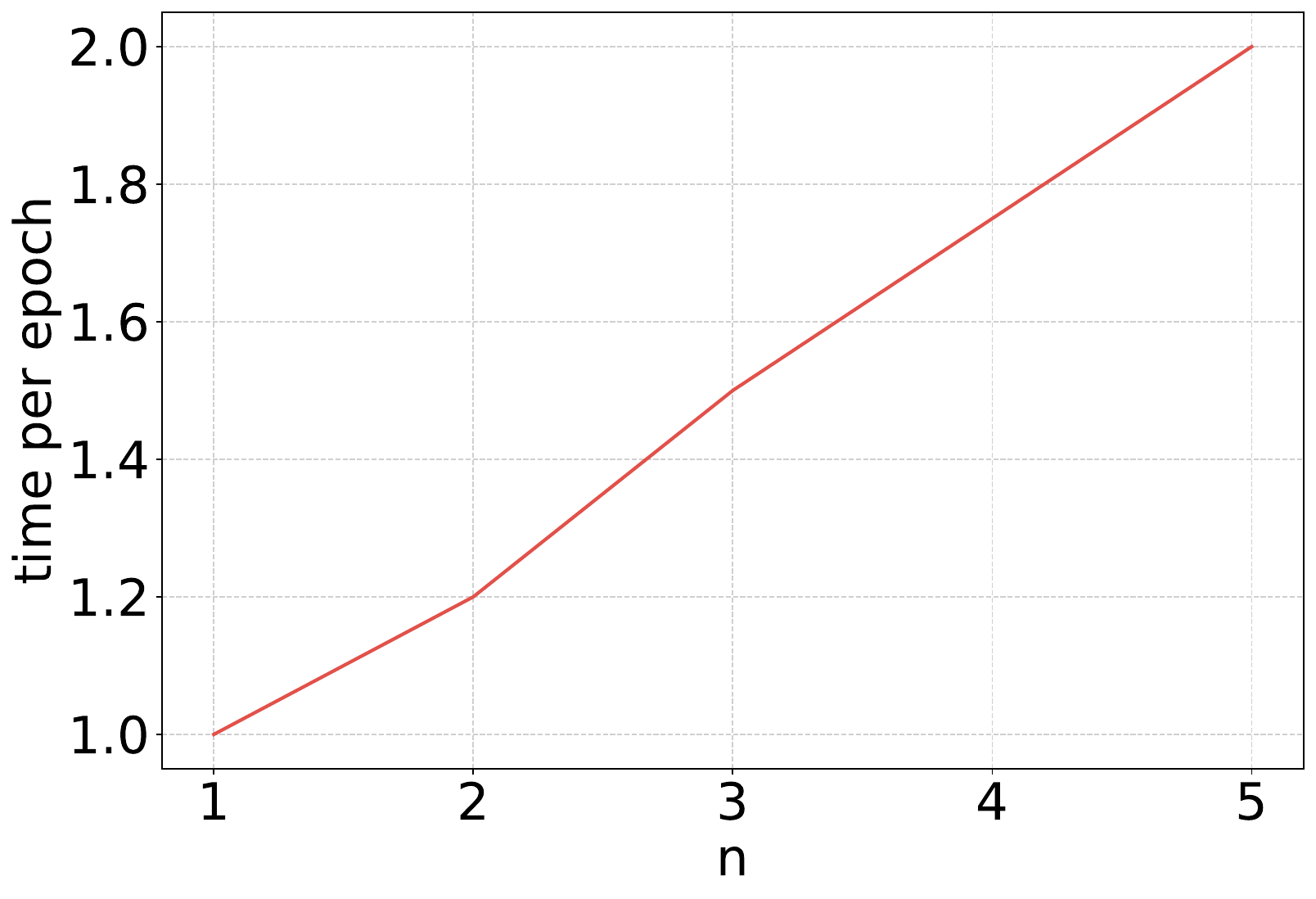}
    \end{subfigure}
\caption{The left panel shows the training error and the maximum norm of the equivalent one-layer SSM computed by $\cref{expansion}$ when using linear SSMs of varying depths, all designed to achieve the same expressivity, to learn a linear functional impulse. The right panel reports the per-epoch runtime of SSMs with different depths, normalized by the runtime of a one-layer linear SSM.}
\label{fig: impluse}
\end{figure}

\paragraph{Experiments on linear functionals}
In this section, we focus on the task of learning a linear functional impulse \citep{jorda2005estimation}, which captures long-range dependencies is known to be challenging for recurrent neural networks (RNNs)\citep{pascanu2013difficulty}. The memory function takes the form of an impulse
\begin{align}
    \rho(s, \alpha) = 
    \begin{cases}
        1 & \text{if } s =\alpha, \\
        0 & \text{otherwise}
    \end{cases}
\end{align}
Here, the parameter $\alpha$ controls the shifting distance. This task is particularly well-suited for highlighting the role of depth in deep linear state-space models, as we keep equal parameter count while increasing depth. The architecture used in these experiments is identical to the one analyzed in our theoretical framework, i.e. deep linear SSMs. We use models of the same effective size, that is, the total width 
$l(m-1)+1$ is fixed. As shown in \cref{fig: impluse}, the approximation error decreases with better performance when we try to fix the same expressivity as the number of layers increases. Each point on the additional line in the plot represents the norm of a one-layer linear SSM that has equivalent representational capacity to the corresponding deep linear SSM. We find that increasing the number of layers while reducing the width by keeping the total number of effective parameters fixed leads to improved performance. However, this improvement comes with a trade-off: as the number of layers increases, the computational speed becomes slower. Furthermore, we observe that as the depth of the model increases, the corresponding norm required by the equivalent one-layer SSM also increases. we compute the corresponding norm of one-layer SSM by \cref{expansion}. This demonstrates that width contributes to improved model performance when the parameter count is fixed.

\begin{figure}[!ht]
    \centering
    \begin{subfigure}[b]{0.45\linewidth}
        \centering
        \includegraphics[width=1\linewidth]{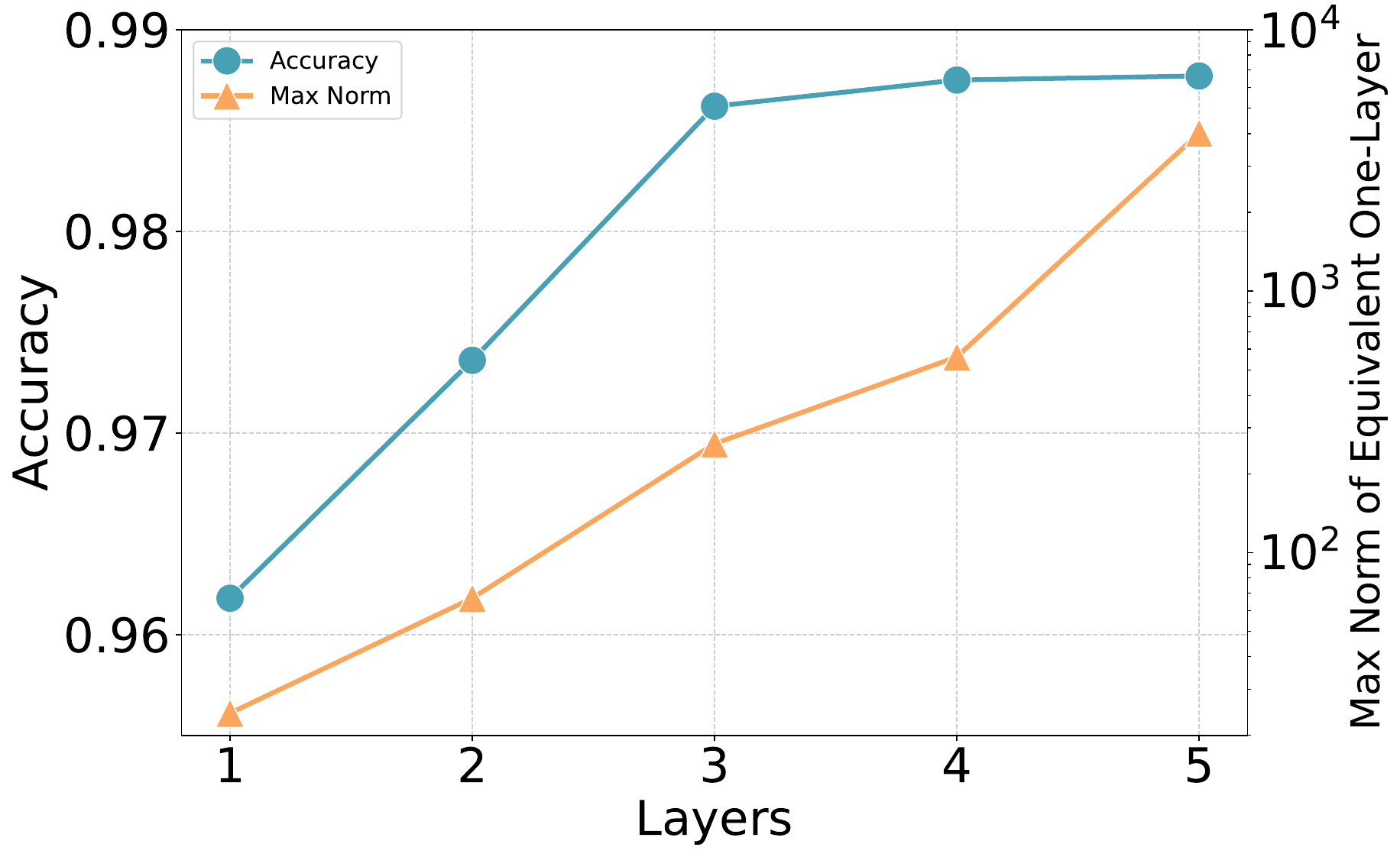}
    \end{subfigure}
    \begin{subfigure}[b]{0.42\linewidth}
        \centering
        \includegraphics[width=1\linewidth]{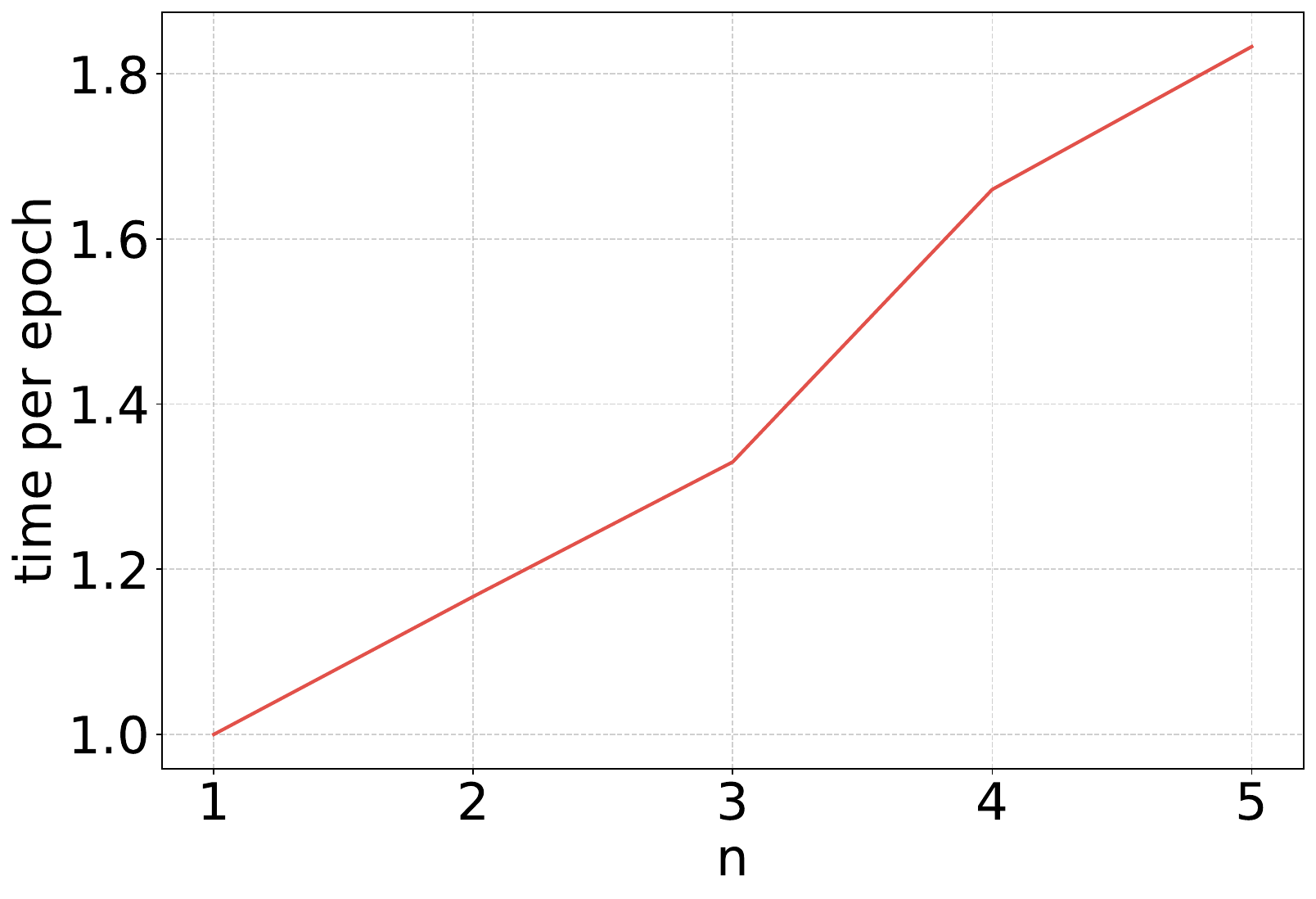}
    \end{subfigure}
\caption{The left panel shows the accuracy and the maximum norm of the equivalent one-layer computed by \cref{expansion} for the S4 model of varying depths on MNIST. All are designed to achieve the same expressivity. The right panel reports the per-epoch runtime of SSMs with different depths, normalized by the runtime of a one-layer SSM.}
\label{fig: mnist}
\end{figure}

\paragraph{Experiments on nonlinear S4 model on MNIST} We aim to validate our theoretical findings on a practical model. To this end, we conduct experiments with the S4 model \citep{smith2022simplified} on the sequential MNIST task \citep{xiao2017fashion}. The architecture used in this experiment is
\begin{align}
     \text{SSM}_{l,m} \to \text{FFN}
\end{align}
where $\text{FFN}$ denotes a feedforward network. We give a clearer explanation in \cref{appendx:S4D structure}. When $l=1$, this reduces to the standard S4D model\citep{gupta2022diagonal}. When $l \geq 1$, we reduce $d_{\text{state}}$ to keep the width fixed, thereby preserving the expressivity of the model. From \cref{fig: mnist}, we observe that the performance improves when a wide one-layer SSM is replaced with a multi-layer, narrower SSM. The experimental results show that increasing the number of layers, while reducing the hidden dimension to maintain the same expressivity, can lead to improved model performance. However, this improvement also comes with the same trade-off: as the number of layers increases, the computational speed becomes slower. Future work may explore the development of more efficient methods for handling multi-layer computations, which could help balance the benefits of depth with practical concerns of computational cost.

\section{Conclusion}
\label{conclusion}
In this work, we investigate the effect of depth on the expressivity of deep linear state-space models under norm constraints. Our theoretical analyses in \cref{theorem 4.1} reveal a nontrivial dependence of model expressivity on depth when the total parameter count is held constant. In particular, while increasing depth and width are generally equivalent in the absence of norm constraints, their roles differ significantly under norm-bounded regimes. In \cref{theorem 4.2}, we also show that increasing depth can significantly reduce the required parameter norms under the same parameter count, especially in tasks that demand large-norm representations such as modeling oscillatory or non-smooth memory functions\citep{pascanu2013difficulty}. This norm reduction highlights the effectiveness of deeper architectures as our experiments shows in \cref{experiment}. Moreover, our results suggest a promising direction for future research: factoring shallow SSMs into deeper ones can enhance expressivity and generalization, even in classical implementations. While this may incur a runtime cost due to increased depth, it opens the door to better optimization and stability, especially when norm control is critical. Understanding this trade-off more thoroughly in nonlinear or real-world SSM variants remains an exciting avenue for future exploration.

There are some important limitations in current work. 
\begin{itemize}
    \item In this paper, we focus on the linear setting, though the results are expected to generalize to the nonlinear case. A key limitation is that the convolutional representation remains unaddressed. Nonetheless, our experiments are conducted on a nonlinear model, which suggests that the theoretical insights may carry over.
    \item In this paper, we analyze the role of depth from an approximation perspective. When considering implicit or explicit regularization dynamics, norm constraints naturally arise, leading to a fundamentally different setting.
\end{itemize}

\bibliography{ref}
\bibliographystyle{ref}

\newpage

\appendix
\section{Proofs of the theorems and lemmas}
\subsection{Proofs on Model Structures and Representations}\label{ProofOfStructure}
\paragraph{Proof of Lemma \ref{lemma: convolution kernel}} It is clear that $h_{n_1}(t)$ is linear w.r.t. $x(s)$, for $n_1 = 1, \dots, l$ and $s = 0, \dots, t$. Then we denote as $h_{(n_1)}(t) = \sum_{s = 0}^{t} h(s, n_1) x(t - s)$.
To prove this lemma, we first apply induction on $s$ and $n_1$ to prove the following statement:
\begin{equation}
    h(s, n_1) = \sum_{\substack{i_{1}+i_{2}+...+i_{n_1}=s\\i_{1},...,i_{n_1}\in\mathbb{N}}}\prod_{j=1}^{n_1}(A_{n_1-j+1}^{i_{n_1-j+1}}B_{n_1-j+1}), \forall n_1 = 1, \dots, l
\end{equation}
When $s = 0$, it is clear that $h(0, n_1) = \prod_{j=1}^{n_1}B_{n_1-j+1}$, $\forall n_1$.
When $n_1 = 1$, it is clear that $h(s, 1) = A_1^{s} B_1$, $\forall s$. 
Then suppose this representation for $h$ is true for $h(s, n_1 - 1)$ and $h(s-1, n_1)$, then we have 
\begin{align}
    h(s, n_1) &= A_{n_1}h(s-1, n_1) + B_{n_1}h(s, n_1 - 1)\\
    &= A_{n_1}\sum_{\substack{i_{1}+i_{2}+...+i_{n_1}=s-1\\i_{1},...,i_{n_1}\in\mathbb{N}}}\prod_{j=1}^{n_1}(A_{n_1-j+1}^{i_{n_1-j+1}}B_{n_1-j+1}) \\ &+ B_{n_1}\sum_{\substack{i_{1}+i_{2}+...+i_{n_1}=s\\i_{1},...,i_{n_1 - 1}\in\mathbb{N}}}\prod_{j=1}^{n_1 - 1}(A_{n_1 - 1-j+1}^{i_{n_1-1-j+1}}B_{n_1-1-j+1})\\
    &= \sum_{\substack{i_{1}+i_{2}+...+i_{n_1}=s\\i_{1},...,i_{n_1}\in\mathbb{N}\\ i_{n_1} \ge 1}}\prod_{j=1}^{n_1}(A_{n_1-j+1}^{i_{n_1-j+1}}B_{n_1-j+1}) \\ &+ \sum_{\substack{i_{1}+i_{2}+...+i_{n_1}=s\\i_{1},...,i_{n_1}\in\mathbb{N}\\ i_{n_1} = 0}}\prod_{j=1}^{n_1}(A_{n_1-j+1}^{i_{n_1-j+1}}B_{n_1-j+1})\\
    &= \sum_{\substack{i_{1}+i_{2}+...+i_{n_1}=s\\i_{1},...,i_{n_1}\in\mathbb{N}}}\prod_{j=1}^{n_1}(A_{n_1-j+1}^{i_{n_1-j+1}}B_{n_1-j+1})
\end{align}
Then we have that with the expression of $h(s-1, n_1)$ and $h(s, n_1 - 1)$, we can have the expression of $h(s, n_1)$ as above. Then with $h(0, n_1)$ and $h(s, 1)$, induction shows the correctness of the above expression. Thus
\begin{equation}
    \rho(t) = C^Th(t, l) = C^T\sum_{\substack{i_{1}+i_{2}+...+i_{l}=t\\i_{1},...,i_{l}\in\mathbb{N}}}\prod_{j=1}^{l}(A_{l-j+1}^{i_{l-j+1}}B_{l-j+1})
\end{equation}
\hfill $\square$

\subsection{Rewriting of Explicit Forms} \label{Rewriting}
Before we come to the proof of \textbf{Theorem \ref{theorem 4.1}}, we prove the following lemmas to aid our proofs. 

\paragraph{Definition 1} Denote by $F_k (\alpha_1, \dots, \alpha_n)$ the following function:
\begin{center}
    $F_k (\alpha_1, \dots, \alpha_n) = \sum_{i = 1}^n \frac{\alpha_i^{k+n-1}}{\prod_{j \ne i} (\alpha_i - \alpha_j)}$
\end{center}
where $n$ is the number of inputs taken into the function, and $\alpha_i$ are distinct complex numbers.
\paragraph{Corollary 1} $F_k (\alpha_1, \dots, \alpha_n) = F_k (\alpha_{\sigma (1)}, \dots, \alpha_{\sigma (n)})$ where $\sigma$ is a permutation. And $F_k (\alpha_1, \dots, \alpha_n) = F_k (\alpha_1, \dots, \alpha_n, 0)$. 
\par The corollary is easy to verify, so we omit the proof here. 
\paragraph{Lemma 3.} For all integer $k$ satisfying $ 0\le k \le n-2$, and distinct complex number $\alpha_1, \dots, \alpha_n$, we have 
\begin{center}
    $\sum_{i=1}^{n} \frac{\alpha_i^{k}}{\prod_{j \ne i} (\alpha_i - \alpha_j)} = 0$
\end{center}
\paragraph{Proof of Lemma 3.} 
For a given integer $1 \le k \le n-2$, consider the following polynomial of $x$ of degree at most $n-1$:
\begin{center}
    $f(x) = \sum_{i=1}^{n} \frac{\alpha_i^{k}\prod_{j \ne i} (x - \alpha_j)}{\prod_{j \ne i} (\alpha_i - \alpha_j)}$
\end{center}
It is obvious by the properties of Lagrange Interpolation Polynomial that $f(x) = x^k$, thus the coefficient of $x^{n-1}$ is $0$, this gives exactly
\begin{center}
    $\sum_{i=1}^{n} \frac{\alpha_i^{k}}{\prod_{j \ne i} (\alpha_i - \alpha_j)} = 0$
\end{center}
\hfill $\square$
\paragraph{Lemma 4.} Denote by $I(n, k)$ the following set
\begin{center}
    $I(n,k) = \{(i_1, \dots, i_n): \sum_{j = 1}^ni_{j} = k; \quad  i_{j} \ge 0, i_{j} \in \mathbb{Z}, \forall j \}$
\end{center}
Then for distinct $\alpha_1, \dots, \alpha_n$, we have
\begin{center}
    $\sum_{(i_1, \dots, i_n) \in I(n,k)} (\prod_{j = 1}^n \alpha_j^{i_j}) = F_k (\alpha_1, \dots, \alpha_n)$
\end{center}
\paragraph{Proof of Lemma 4.} We do induction over $n$. It is easy to calculate that when $n = 1$, $\forall k \ge 0$, 
\begin{center}
    $\sum_{(i_1, \dots, i_n) \in I(n,k)} (\prod_{j = 1}^n \alpha_j^{i_j}) = \alpha_1^k = F_k (\alpha_1)$
\end{center}
Now suppose that for $n-1$ and $\forall k \ge 0$, the equalities holds. Then we consider the case of $n$. 
\begin{align}
    \sum_{(i_1, \dots, i_n) \in I(n,k)} (\prod_{j = 1}^n \alpha_j^{i_j}) &= \sum_{k_1 = 0}^k \alpha_{n}^{k - k_1}(\sum_{(i_1, \dots, i_{n-1}) \in I(n-1,k_1)} (\prod_{j = 1}^{n-1} \alpha_j^{i_j})) \\
    &= \sum_{k_1 = 0}^k \alpha_{n}^{k - k_1}F_{k_1}(\alpha_1, \dots, \alpha_{n-1})\\
    &= \sum_{k_1 = 0}^k  \sum_{i = 1}^{n-1} \frac{\alpha_i^{k_1+n-2}\alpha_{n}^{k - k_1}}{\prod_{j \ne i;j \ne n} (\alpha_i - \alpha_j)}\\
    &= \sum_{i = 1}^{n-1} \frac{\alpha_i^{k+n-1} - \alpha_n^{k+1}\alpha_i^{n-2}}{(\prod_{j \ne i; j \ne n} (\alpha_i - \alpha_j)) (\alpha_i - \alpha_n)}\\
    &= \sum_{i = 1}^{n-1} \frac{\alpha_i^{k+n-1} - \alpha_n^{k+1}\alpha_i^{n-2}}{\prod_{j \ne i} (\alpha_i - \alpha_j)}\\
    &= \sum_{i = 1}^{n-1} \frac{\alpha_i^{k+n-1}}{\prod_{j \ne i} (\alpha_i - \alpha_j)} + \alpha_n^{k+1}(0 - \sum_{i = 1}^{n-1} \frac{\alpha_i^{n-2}}{\prod_{j \ne i} (\alpha_i - \alpha_j)})\\
    &= \sum_{i = 1}^{n-1} \frac{\alpha_i^{k+n-1}}{\prod_{j \ne i} (\alpha_i - \alpha_j)} + \alpha_n^{k+1} (\frac{\alpha_n^{n-2}}{\prod_{j \ne n} (\alpha_n - \alpha_j)})\\
    &= F_k(\alpha_1, \dots, \alpha_n)
\end{align}
Where the second last step uses the result of \textbf{Lemma 3}. 
\hfill $\square$
\par As the set $\{(\alpha_1, \dots, \alpha_n): \text{$\alpha_1, \dots, \alpha_n$ are distinct}\}$ is open and dense in $\mathbb{C}^n$, and taking finite sum in the form of $\sum_{(i_1, \dots, i_n) \in I(n,k)} (\prod_{j = 1}^n \alpha_j^{i_j}) = F_k (\alpha_1, \dots, \alpha_n)$ is continuous w.r.t. $(\alpha_1, \dots, \alpha_n)$ under norms on $\mathbb{C}^n$, thus we can extend the definition of $F_k(\alpha_1, \dots, \alpha_n)$ to the entire $\mathbb{C}^n$ by taking limits. 
\paragraph{Corollary 2.} For a $l$ layer SSM $\rho \in \mathcal{H}_{\infty,l}^{m}$, suppose it is defined by matrix $A_i, i = 1, \dots, l$; $B_i, i = 1, \dots, l$ and $C$ as given in the formulation \ref{linear deep ssm}. Then we have 
\begin{equation}
    \rho(t)= \sum_{1\le j_{1},...,j_{l}\le m}C_{j_l}[\prod_{p=1}^{l-1}(B_{l-p+1})_{(j_{l-p+1}, j_{l-p})}] (B_1)_{j_1}[ \sum_{\substack{i_{1}+i_{2}+...+i_{l}=t\\i_{1},...,i_{l}\in\mathbb{N}}}\prod_{j=1}^{l}((A_{l-j+1})_{(j_{l-p+1}, j_{l-p+1})}^{i_{l-j+1}}) ]
\end{equation}
If we denote $\alpha(x,y)$ as $\alpha(x,y) = (A_x)_{(y,y)}$ for diagonal matrices $A_x$, then we have 
\begin{equation}
    \rho(t)= \sum_{1\le j_{1},...,j_{l}\le m}C_{j_l}[\prod_{p=1}^{l-1}(B_{l-p+1})_{(j_{l-p+1}, j_{l-p})}] (B_1)_{j_1}F_t(\alpha(1, j_1), \dots, \alpha(l, j_l))
\end{equation}
This corollary is straight from calculations of \textbf{Lemma \ref{lemma: convolution kernel}}
\paragraph{Lemma 5.} Given fixed positive integers $M \ge N, T > 0$. Denote by $B(\alpha, \epsilon) = \{c \in \mathbb{C}: |c-\alpha| < \epsilon\}$ for real number $\epsilon > 0$. For fixed complex numbers $b_1, \dots, b_N$ satisfying $\sum_{i = 1}^{N} b_i \ne 0$, and pairwise distinct non-zero complex numbers $\alpha_1, \dots, \alpha_n$. Then for $T > N + M + 1$ and sufficiently small $\epsilon$, we have 
\begin{equation}
    \lim_{\epsilon \to 0}(\inf_{\substack{\beta_1, \dots, \beta_M \in \cup_{i = 1}^N B(\alpha_i, \epsilon)\\ \sum_{i = 1}^M c_i = 0}} \sum_{t = 1} ^ T |(\sum_{i=1}^N b_i \alpha_i^t)-(\sum_{i = 1}^M c_i \beta_i^t)|) > f(\{\alpha_i\}, \{b_i\}) >0
\end{equation}
Where $f(\{\alpha_i\}, \{b_i\})$ is a real number independent of $\epsilon$. 
\paragraph{Proof of Lemma 5.} It is clear that for sufficiently small $\epsilon$, $\beta_i$ have to be non-zero, and $B(\alpha_i, \epsilon)$ are disjoint. Then we rewrite $(\sum_{i = 1}^M c_i \beta_i^t)$ as 
\begin{equation}
    \sum_{i = 1}^M c_i \beta_i^t = \sum_{j = 1}^N \sum_{\beta_{ji} \in B(\alpha_j, \epsilon)} c_{ji} \beta_{ji}^t
\end{equation}
Then by considering the first-order asymptotic of $\sum_{\beta_{ji} \in B(\alpha_j, \epsilon)} c_{ji} \beta_{ji}^t$ w.r.t. $\epsilon$, we have that 
\begin{equation}
    \sum_{\beta_{ji} \in B(\alpha_j, \epsilon)} c_{ji} \beta_{ji}^t = B_j(t) \alpha_j^t + o(1) 
\end{equation}
Where $B_j(t)$ are polynomials with $\sum_{j = 1}^N deg(B_j) \le M - N$ (The possible existence of polynomial comes from cases where $\sum_{\beta_{ji} \in B(\alpha_j, \epsilon)} c_{ji}= 0$, leading to cancellation of the originally highest order term. In the first order asymptotic we consider the highest order not canceled out in the summation). 
\par If $\sum_{j = 1}^N deg(B_j) \ne 0$, then it is clear that such $f(\{\alpha_i\}, \{b_i\})$ exists, as at least one $B_j(t)$ grows at least linearly w.r.t. $t$ but the corresponding $b_j$ remains constant. 
\par If $\sum_{j = 1}^N deg(B_j) = 0$, then there's no highest-order cancellation in the way above (canceling the entire term doesn't count in this case). Then by writing $B_j(t) = B_j$, then minium becomes 
\begin{equation}
    \lim_{\epsilon \to 0}(\inf_{\sum_{i = 1}^M B_i = 0} \sum_{t = 1} ^ T |(\sum_{i=1}^N b_i \alpha_i^t)-(\sum_{i = 1}^N B_i \alpha_i^t)| + O(\epsilon))
\end{equation}
And it is clear from $\sum_{i = 1}^N b_i \ne 0$ that such limit is strictly greater than zero. Thus the $f(\{\alpha_i\}, \{b_i\})$ exists. 
\hfill $\square$

\subsection{Proofs of Main Theorems} \label{ProofMain}
\paragraph{Lemma 6.} $\mathcal{H}_{\infty,1}^{l(m-1)+2}\not\subseteq\mathcal{H}_{\infty,l}^{m}$. 
\paragraph{Proof of Lemma 6.} We consider $\rho \in \mathcal{H}_{\infty,1}^{l(m-1)+2}$ defined by matrices $A_0, B_0, C_0$, where $A_0 = Diag\{\sigma_1, \dots, \sigma_K\}$ with $K = l(m-1) +2$ and distinct nonzero $\sigma_i$'s. Then we have $\rho(t) = \sum_{i=1}^K (B_0)_i(C_0)_i \sigma_i^t$. 
Assume the contrary, if $\rho(t) \in \mathcal{H}_{\infty,l}^{m}$, then suppose it is defined by $A_1, \dots, A_l$; $B_1, \dots, B_l$; $C$ as in the formulation, and denote $\alpha(x, y) = (A_x)_{(y,y)}$ to be the diagonal elements of $A_x$. According to the formulation in \textbf{Corollary 2.}, it is clear that $\{\alpha(i,j): 1 \le i \le l, 1\le j \le m\} = \{\sigma_i: 1 \le i \le K\} \cup \{0\}$.\\
\par If there are any two non-zero $\alpha(x_1,y_1) = \alpha(x_2, y_2)$, then we perturb them up to the magnitude of $\epsilon$ such that all the non-zero $\alpha(x, y)$ are distinct, and satisfying that $\alpha(x, y) \in (\cup_{i = 1}^K B(\sigma_i, \epsilon)) \cup\{0\}$. As $\epsilon \to 0$, the perturbation of $\rho(t)$ also goes to $0$ for any fixed $t$.  

\par Now we consider $F_t(\alpha(1, j_1), \dots, \alpha(l, j_l))$, and eliminate all the zeros according to the rule in \textbf{Corollary 1.}. Then we denote that $F_t(\alpha(1, j_1), \dots, \alpha(l, j_l)) = F_t(\beta_1, \dots, \beta_{l_1})$, where $l_1 \ge 2$ due to that there's at most $l-2$ zeros, and $\beta_i$ are all the non-zero elements in $\{\alpha(i, j_i)\}$, and $\beta_i$ are distinct due to that non-zero $\{\alpha(i, j_i)\}$ are perturbed to be distinct. 
\par Recalling the definition, we have
\begin{align}
    F_t(\beta_1, \dots, \beta_{l_1}) &= \sum_{i = 1}^{l_1} \frac{\beta_i^{t+l_1-1}}{\prod_{j \ne i} (\beta_i - \beta_j)}\\
    &= \sum_{i = 1}^{l_1} (\frac{\beta_i^{l_1-2}}{\prod_{j \ne i} (\beta_i - \beta_j)}) \beta_i^{t+1}
\end{align}
where the sum of coefficients is 
\begin{equation}
    \sum_{i = 1}^{l_1} (\frac{\beta_i^{l_1-2}}{\prod_{j \ne i} (\beta_i - \beta_j)}) = 0
\end{equation}
As according to \textbf{Lemma 3}. And this applies for all such$F_t(\alpha(1, j_1), \dots, \alpha(l, j_l))$. Then by taking $N = K$, $M = lm$, $b_i = \frac{(B_0)_i(C_0)_i}{\sigma_i}$, $\alpha_i = \sigma_i$, and $\beta_j = \alpha(x,y)$ (with the perturbation applied), then we have exactly the setting of \textbf{Lemma 5}. 
\par Then according to \textbf{Lemma 5}, we have that for the perturbed $\hat{\rho}(t)$, $\sum_{t = 1}^{2ml+1} |\rho(t) - \hat{\rho}(t)| > f_0 > 0$, where $f_0$ is completely determined by the original system defining $\rho$ and independent of the magnitude of perturbation $\epsilon$. 
\par However, we know that for any bounded range of $t$, the summation $\sum_{\substack{i_{1}+i_{2}+...+i_{l}=t\\i_{1},...,i_{l}\in\mathbb{N}}}\prod_{j=1}^{l}((A_{l-j+1})_{(j_{l-p+1}, j_{l-p+1})}^{i_{l-j+1}})$ in \textbf{Corollary 2} has perturbation bounded by a function of $\epsilon$ that goes to zero for $\epsilon \to 0$. This is in contradiction with the positive bound $f_0$. Thus concludes the proof. 
\hfill $\square$
\paragraph{Lemma 7.} $\mathcal{H}_{\infty,l}^{m}\subseteq\mathcal{H}_{\infty,1}^{lm}$.
\paragraph{Proof of Lemma 7.} For $\rho \in \mathcal{H}_{\infty,l}^{m}$ defined by matrices $A_i, i = 1, \dots, l$; $B_i, i = 1, \dots, l$; $C$ according to the formulation, we construct a $1$ layer SSM of width $ml$ defined by vector $C_0, B_0 \in \mathbb{C}^{ml \times 1}$ and $A_0 \in \mathbb{C}^{ml\times ml}$, such that the hidden state $f^{(t)}$ at time $t$ is exactly $({h}^T_1(t), \dots, {h}^T_l(t))^T$. 
\par We denote $A_0 = (A_0)_{(ij)}, 1 \le i, j\le l$, where each $(A_0)_{(ij)}$ is a $m \times m$ block. And $B_0 = (\beta_1, \dots, \beta_l)$, $C_0 = (c_1, \dots, c_l)$ are separated into $l$ of $m \times 1$ vectors stacked together.
\begin{equation}
    (A_0)_{(ij)}=\left\{\begin{matrix}
 (\prod_{k = 1}^{i-j} B_{i +1 - k}) A_j, i > j\\
 A_i, i = j\\
0, i < j
\end{matrix}\right.
\end{equation}
And $b_i = \prod_{j = 1}^{i} B_{i+1 -j}$, $C_0 = (0, \dots, 0, C)$. By direct calculation, we could verify that the $\hat\rho$ corresponding to the SSM defined by $A_0, B_0, C_0$ is exactly $\rho$, which means $\rho \in \mathcal{H}_{\infty,1}^{lm}$. 
\hfill $\square$
\paragraph{Corollary 3.} For distinct complex numbers $\gamma_1, \dots, \gamma_n$, we have that for non-negative integer $t$ 
\begin{equation}
    F_t(\gamma_1, \dots, \gamma_n) - \frac{\gamma_{n-1}}{\gamma_{n-1} - \gamma_{n}}F_t(\gamma_1, \dots, \gamma_{n-1}) = \frac{\gamma_n}{\gamma_n - \gamma_{n-1}}F_t(\gamma_1, \dots, \gamma_{n-2}, \gamma_{n})
\end{equation}
This result in \textbf{Corollary 3} could be derived directly by expanding $F_t$ according to \textbf{Definition 1} and simple calculations. 
\paragraph{Lemma 8.} Given $n$ distinct non-zero complex number $\beta_1, \dots, \beta_n$, and $n$ complex number $Z_1, \dots, Z_n$. If we define $H_k, k = 1, \dots, n$ to be
\begin{equation}
    H_k = \sum_{j = k}^n Z_j \frac{\beta_k\prod_{p = 1}^{k - 1}(\beta_j - \beta_p)}{\beta_j^{k}}
\end{equation}
Then we have 
\begin{equation}
    \sum_{k = 1}^n H_k F_t(\beta_1, \dots, \beta_k) = \sum_{k =1}^n Z_k \beta_k^t
\end{equation}
Furthermore, if $|\beta_i|$ is in an non-decreasing order, then we have $|H_k| \le 2^n \max_{1\le i \le n} |Z_i|, \forall k$. 
\paragraph{Proof of Lemma 8.} We use induction to prove the following stronger result for all $u = 1, \dots, n$:
\begin{equation}
    \sum_{k = u}^n H_k F_t(\beta_1, \dots, \beta_k) = \sum_{k = u}^{n} Z_k \frac{\prod_{p =1}^{u-1}(\beta_k - \beta_p)}{\beta_k^{u-1}}F_t(\beta_1, \dots, \beta_{u-1}, \beta_k)
\end{equation}
In particular, when $u = 1$, it becomes exactly the result we want. 
\par When $u = n$, it is clear that the above is true. We then calculate the following for $u \ge 2$:
\begin{align}
    &\sum_{k = u-1}^{n} Z_k \frac{\prod_{p =1}^{u-2}(\beta_k - \beta_p)}{\beta_k^{u-2}}F_t(\beta_1, \dots, \beta_{u-2}, \beta_k) - \sum_{k = u}^{n} Z_k \frac{\prod_{p =1}^{u-1}(\beta_k - \beta_p)}{\beta_k^{u-1}}F_t(\beta_1, \dots, \beta_{u-1}, \beta_k)\\
    &= Z_{u-1} \frac{\prod_{p =1}^{u-2}(\beta_{u-1} - \beta_p)}{\beta_{u-1}^{u-2}}F_t(\beta_1, \dots, \beta_{u-2}, \beta_{u-1}) \\
    &+ \sum_{k = u}^{n} Z_k (\frac{\prod_{p =1}^{u-2}(\beta_k - \beta_p)}{\beta_k^{u-2}}F_t(\beta_1, \dots, \beta_{u-2}, \beta_k)-\frac{\prod_{p =1}^{u-1}(\beta_k - \beta_p)}{\beta_k^{u-1}}F_t(\beta_1, \dots, \beta_{u-1}, \beta_k))\\
    &= Z_{u-1} \frac{\prod_{p =1}^{u-2}(\beta_{u-1} - \beta_p)}{\beta_{u-1}^{u-2}}F_t(\beta_1, \dots, \beta_{u-2}, \beta_{u-1}) \\
    &+ \sum_{k = u}^{n} Z_k \frac{\prod_{p =1}^{u-1}(\beta_k - \beta_p)}{\beta_k^{u-1}}(\frac{\beta_k}{(\beta_k - \beta_{u-1})}F_t(\beta_1, \dots, \beta_{u-2}, \beta_k)-F_t(\beta_1, \dots, \beta_{u-1}, \beta_k))\\
    &= Z_{u-1} \frac{\prod_{p =1}^{u-2}(\beta_{u-1} - \beta_p)}{\beta_{u-1}^{u-2}}F_t(\beta_1, \dots, \beta_{u-2}, \beta_{u-1}) \\
    &+ \sum_{k = u}^{n} Z_k \frac{\prod_{p =1}^{u-1}(\beta_k - \beta_p)}{\beta_k^{u-1}}(\frac{\beta_{u-1}}{\beta_{k} - \beta_{u-1}}F_t(\beta_1, \dots,\beta_{u-2} , \beta_{u-1}))\\
    &= \sum_{k = u-1}^{n} Z_k \frac{\beta_{u-1}\prod_{p =1}^{u-2}(\beta_k - \beta_p)}{\beta_k^{u-1}}(F_t(\beta_1, \dots,\beta_{u-2} , \beta_{u-1}))\\
    &= H_{u-1}F_t(\beta_1, \dots,\beta_{u-2} , \beta_{u-1})
\end{align}
Then it is clear that the stronger result holds for all $u = 1, \dots, n$. Thus taking $u = 1$ we have the desired result. 
\par If the non-decreasing order of $|\beta_i|$ holds, then we have that 
\begin{align}
    |H_k| &= |\sum_{j = k}^n Z_j \frac{\beta_k\prod_{p = 1}^{k - 1}(\beta_j - \beta_p)}{\beta_j^{k}}| \\
    &\le \sum_{j = k}^n |Z_j| *|\frac{\beta_k}{\beta_j}|*{\prod_{p = 1}^{k - 1}|\frac{(\beta_j - \beta_p)}{\beta_j}|}\\
    &\le \sum_{j = k}^n |Z_j|*2^{k-1}\\
    &\le 2^n \max_{1\le i \le n} |Z_i|
\end{align}
\hfill $\square$
\paragraph{Lemma 9.} $\mathcal{H}_{\infty,1}^{l(m-1)+1}\subseteq\mathcal{H}_{\infty,l}^{m}$.
\paragraph{Proof of Lemma 9.} For simplicity we denote $K = l(m-1)+1$. 
\par Then for $\rho \in \mathcal{H}_{\infty,1}^{l(m-1)+1}$, suppose that it is defined by vectors $B_0, C_0$, and matrix $A_0 = Diag\{\sigma_1, \dots, \sigma_K\}$. Then from \textbf{Corollary 2} we have $\rho(t) = \sum_{i =1}^K (C_0)_i (B_0)_i \sigma_i^t$.  
\par As the case of $\sigma_i = \sigma_j$ or $\sigma_i = 0$ will lead to the degenerate case, which is weaker than the non-degenerate case, thus we do not consider them here. Now we assume that $\sigma_i$ are distinct and non-zero. Due to the symmetry of the above form, we can assume without loss of generality that $|\sigma_i|$ is in an non-decreasing order. 
\par we define the following functions:
\begin{equation}
    \alpha(i,j) = \left\{\begin{matrix}
 \sigma_j, \quad \text{for }i = 1; j = 1, \dots, m  \\
 \sigma_{(i-1)(m-1)+ j +1 }, \quad \text{for }i = 2, \dots, l; j = 1, \dots, m\\
0, \quad \text{for }i = 2, \dots, l; j = m
\end{matrix}\right.
\end{equation}
\begin{equation}
    Z(i,j) = \left\{\begin{matrix}
 (B_0)_j(C_0)_j, \quad \text{for }i = 1; j = 1, \dots, m  \\
 (B_0)_{(i-1)(m-1)+ j +1 }(C_0)_{(i-1)(m-1)+ j +1 }, \quad \text{for }i = 2, \dots, l; j = 1, \dots, m\\
0, \quad \text{for }i = 2, \dots, l; j = m
\end{matrix}\right.
\end{equation}
\begin{equation}
    H(i,j) = \left\{\begin{matrix}
 \sum_{p = i}^l Z(p, j) \frac{\alpha(i,j)\prod_{q = 1}^{i - 1}(\alpha(p.j) - \alpha(q, j))}{\alpha(p, j)^{i}}, \quad \text{for } i = 1, \dots, l; j = 1, \dots, m-1\\
 Z(1, m), \quad \text{for } i = 1; j = m\\
0, \quad \text{for }i = 2, \dots, l; j = m
\end{matrix}\right.
\end{equation}
Also, denote by $Z_0 = 2 *(\max_{1 \le i \le K} |(B_0)_i(C_0)_i|)^{\frac{1}{l+1}}$. 
\par Then we construct diagonal matrices $A_1, \dots, A_l$; matrices $B_2, \dots, B_l$; vectors $B_1, C$ such that the $l$ layer SSM of width $m$ defined by these parameters has exactly the same $\rho$. 
\begin{itemize}
    \item $B_0 = C = Z_0 1_m$, where $1_m$ is the vector consisting of all $1$.
    \item $(A_i)_{(j,j)} = \alpha(i,j)$, for $i = 1, \dots, l; j =1, \dots, m $.
    \item $(B_i)_{(j,j)} = Z_0$, for $i = 2, \dots, l-1; j = 1, \dots, m-1$. 
    \item $(B_i)_{(m,m)} = Z_0$, for $i = 3, \dots, l$.
    \item $(B_2)_{(m,m)} = \frac{H(1,m)}{Z_0^l}$. 
    \item $(B_l)_{(j,j)} = \frac{H(l,j)}{Z_0^l}$, for $j = 1, \dots, m-1$.
    \item $(B_i)_{(m, j)} = \frac{H(i-1,j)}{Z_0^l}$, for $i = 2, \dots, l; j = 1, \dots, m-1$
    \item All the unmentioned elements are set to zero.
\end{itemize}
Under such construction, we suppose it defines $\hat\rho$, then following \textbf{Corollary 2} and \textbf{Lemma 8}, we have 
\begin{align}
    \hat\rho(t) &= \sum_{1\le j_{1},...,j_{l}\le m}C_{j_l}[\prod_{p=1}^{l-1}(B_{l-p+1})_{(j_{l-p+1}, j_{l-p})}] (B_1)_{j_1}F_t(\alpha(1, j_1), \dots, \alpha(l, j_l))\\
    &= Z(1,m)\alpha(1,m)^t + \sum_{j = 1}^{m-1}\sum_{i = 1}^l H(i,j)F_t(\alpha(1,j), \dots, \alpha(i,j))\\
    &= Z(1,m)\alpha(1,m)^t + \sum_{j = 1}^{m-1}\sum_{i = 1}^l Z(i,j)\alpha(i,j)^t\\
    &= \sum_{i =1}^K (C_0)_i (B_0)_i \sigma_i^t\\
    &= \rho(t)
\end{align}
Thus $\rho = \hat\rho \in \mathcal{H}_{\infty,l}^{m}$. 
\hfill $\square$
\paragraph{Remark 1.}: As for the degenerated cases, as there are less parameters in $\rho_t$, its construction could be achieved by simple modifications on the above construction, or simply achieved by taking limits as the set of non-degenerate $(\sigma_1, \dots, \sigma_K)$ is open and dense in $\mathbb{C}^{K}$. 
\paragraph{Remark 2.}: It can be seen from \textbf{Lemma 8} that all the entries of $B_i$, $C$ are bounded by $Z_0 = 2 *(\max_{1 \le i \le K} |(B_0)_i(C_0)_i|)^{\frac{1}{l+1}}$. As for the degenerated cases, we can do the construction by taking limits, thus for these cases the same bound holds. 

\paragraph{Proof of Theorem \ref{theorem 4.1}} This is direct from \textbf{Lemma 6}, \textbf{Lemma 7}, \textbf{Lemma 9}. 
\hfill $\square$
\paragraph{Proof of Theorem \ref{theorem 4.2}} For $\rho\in\mathcal{H}_{c_{1},1}^{l(m-1)+1}$, we have that 
\begin{equation}
    Z_0 = 2 *(\max_{1 \le i \le K} |(B_0)_i(C_0)_i|)^{\frac{1}{l+1}} \le 2c_1^{\frac{2}{l+1}}
\end{equation}
Then it is direct from \textbf{Lemma 9, Remark 2}.
\hfill $\square$
\paragraph{Proof of Theorem \ref{theorem 4.3}} From \textbf{Theorem \ref{theorem 4.2}}, it suffices that 
\begin{equation}
    2c_1^{\frac{2}{l+1}} \le c_2
\end{equation}
Which means 
\begin{equation}
    l \ge \lceil\frac{2\ln(c_{1})}{\ln{(\frac{c_{2}}{2})}}-1\rceil
\end{equation}
For this $l$, we already have that $m = \lceil \frac{K}{l} \rceil +1$ satisfies that $K \le l(m-1)$, and \textbf{Lemma 9, Remark 2} shows that the bound also holds for degenerate cases, which is valid to be used here. 
\hfill $\square$
\paragraph{Proof of Corollary \ref{Hermite property}} To prove this corollary, we only need to show that $\mathcal{G}_{c_{1},1}^{l(m-1)+1} \subseteq \mathcal{H}_{\sqrt{l(m-1)+1}c_{1},1}^{l(m-1)+1}$. Assume $\rho \in \mathcal{G}_{c_{1},1}^{l(m-1)+1}$ is defined by $A_1, B_1, C$, and based on the property of normal matrices, we can decompose as $A_1 = U\Sigma U^*$ for diagonal $\Sigma$ and unitary $U$. Then we have 
\begin{align}
    \rho(t) &= C^T A_1^{t}B_1\\
    &= C^T U \Sigma^t U^*B_1 \\
    &= (U^TC)^T \Sigma^t (U^*B_1)
\end{align}
Therefore $\rho$ is also defined by $\Sigma, (U^TC), (U^* B_1)$. And by the norm-preserving property of unitary matrix, we have $\|(U^TC)\|_2 = \|C\|_2$, $\|U^*B_1\|_2 \le \|B_1\|_2$. Then $\|(U^TC)\|_\infty \le \sqrt{l(m-1)+1}c_{1}$, $\|(U^*B_1)\|_\infty \le \sqrt{l(m-1)+1}c_{1}$, leading to $\rho \in \mathcal{G}_{\sqrt{l(m-1)+1}c_{1},1}^{l(m-1)+1}$. Then by \textbf{Theorem \ref{theorem 4.2}} , the conclusion holds. 
\hfill $\square$
\paragraph{Proof of Lemma \ref{expansion}} This lemma can be directly obtained from \textbf{Corollary 2} by indexing $F_t(\alpha(1, j_1), \dots, \alpha(l, j_l))$ as according to \textbf{Definition 1}. 
\hfill $\square$

\paragraph{Remark 3.} We could define an even larger space as follows: 
\begin{align}
    \mathcal{L}_{c,l}^{m}=&\{\rho(t):y(t)=(\rho\ast x)(t),A_{1},...,A_{l}\in\mathbb{C}^{m\times m}\thinspace\mbox{diagonalizable},B_{2},...,B_{l},\in\mathbb{C}^{m\times m},\\ &C,B_{1}\in\mathbb{C}^{m\times1},\max_{i=1,...,l}r(A_{i})<1,||C||_{\infty}\le c, ||B_{1}||_{\infty}\le c,\max_{2\le k\le l}\max_{1\le i,j\le m}|(B_{k})_{ij}|\le c\}
\end{align}

The difference between $\mathcal{H}_{c,l}^{m}$ and $\mathcal{L}_{c,l}^{m}$ is not as substantial as one might initially expect. In fact, if we remove the norm constraint-that is, as $c$ approaches infinity-the two hypothesis spaces become identical.
\begin{lemma}[Representing ability equivalence in two hypothesis space]
\label{lemma: equaivalence}
Given $m,l\ge1$, the we have 
\begin{equation}
    \mathcal{L}_{\infty,l}^{m}=\mathcal{H}_{\infty,l}^{m}
\end{equation}
\end{lemma}

\paragraph{Proof of Lemma \ref{lemma: equaivalence}} As it is clear that $$\mathcal{H}_{\infty,l}^{m} \subseteq \mathcal{L}_{\infty,l}^{m}$$. We only need to prove that $\forall \rho \in \mathcal{L}_{\infty,l}^{m}$, we have $ \rho \in \mathcal{H}_{\infty,l}^{m}$. 
Suppose this $l$ layer SSM is defined by matrix $A_i, i = 1, \dots, l$; $B_i, i = 1, \dots, l$ and $C$ as given in the formulation. 
Then suppose $A_i = P_i^{-1} D_i P_i$ is the Jordan decomposition of $A_i$, where $P_i$ is invertible and $D_i$ is diagonal. 
Then let $\hat{C} =  P_l^T C$, $\hat{B}_j = P_j^{-1} B_jP_{j-1}, j = 2, \dots, l$, $\hat{B}_1 = P_1^{-1}B_1$, $\hat{A}_i = D_i, i = 1, \dots, l$. It is clear that for all $t$, 
\begin{align}
    \rho(t) &= C^T\sum_{\substack{i_{1}+i_{2}+...+i_{l}=t\\i_{1},...,i_{l}\in\mathbb{N}}}\prod_{j=1}^{l}(A_{l-j+1}^{i_{l-j+1}}B_{l-j+1}) \\
    &= \hat{C^T}\sum_{\substack{i_{1}+i_{2}+...+i_{l}=t\\i_{1},...,i_{l}\in\mathbb{N}}}\prod_{j=1}^{l}(\hat{D}_{l-j+1}^{i_{l-j+1}}\hat{B}_{l-j+1})
\end{align}
As $D_i$ are diagonal matrices, we have that $\rho \in \mathcal{H}_{\infty,l}^{m}$. 
\hfill $\square$

\section{Experiment Details}
The model architecture used in our experiments on the nonlinear S4 model for the MNIST dataset is described as follows:
\label{appendx:S4D structure}
\begin{figure}[!ht]
   \centering
   \includegraphics[width=0.5\textwidth]{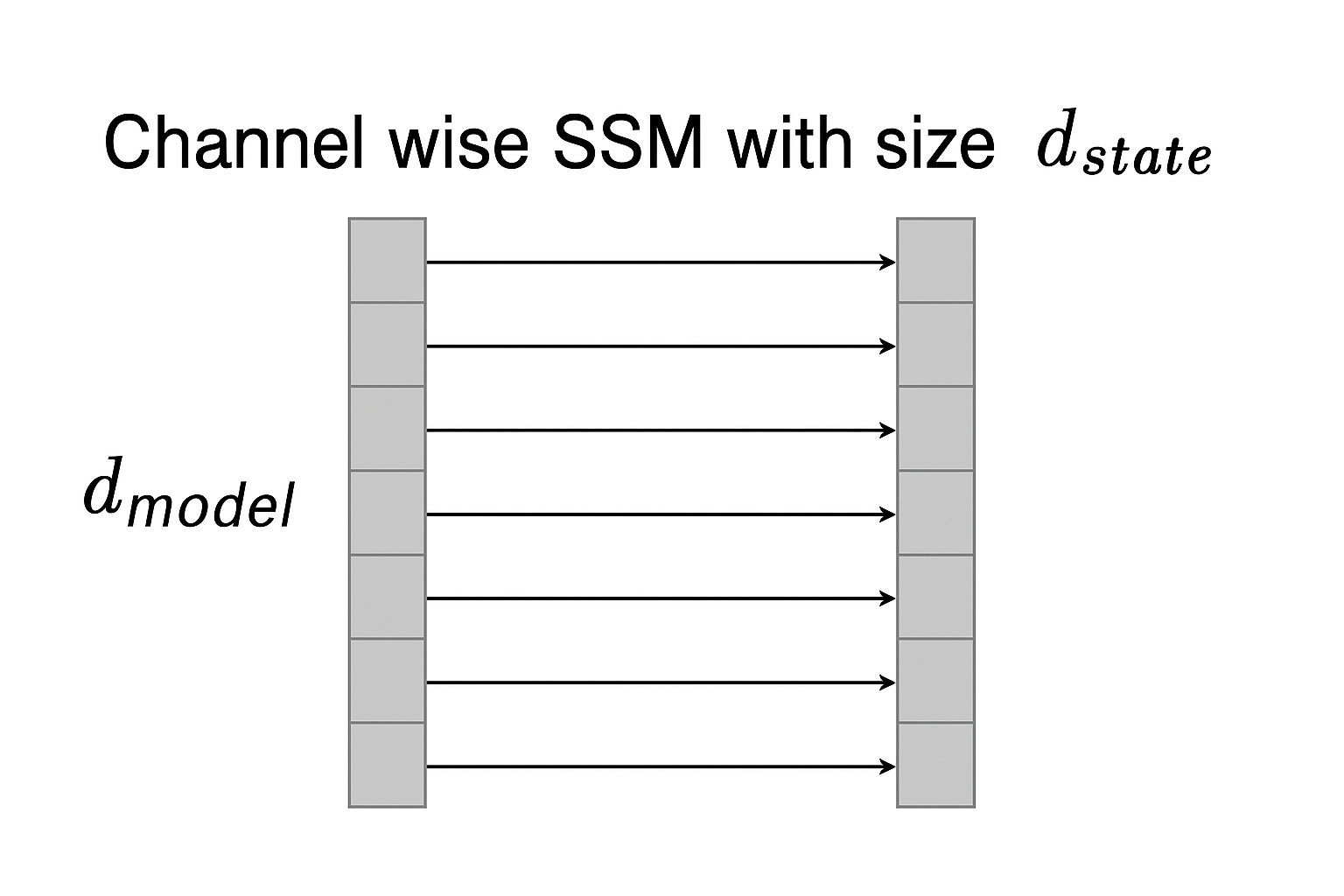}
   \label{fig: S4D structure}
\end{figure}

\newpage
\section*{NeurIPS Paper Checklist}

\begin{enumerate}

\item {\bf Claims}
    \item[] Question: Do the main claims made in the abstract and introduction accurately reflect the paper's contributions and scope?
    \item[] Answer: \answerYes{} 
    \item[] Justification: The contributions and scope are clearly stated in the abstract and introduction.

\item {\bf Limitations}
    \item[] Question: Does the paper discuss the limitations of the work performed by the authors?
    \item[] Answer: \answerYes{} 
    \item[] Justification: The limitations is discussed in \cref{conclusion}

\item {\bf Theory assumptions and proofs}
    \item[] Question: For each theoretical result, does the paper provide the full set of assumptions and a complete (and correct) proof?
    \item[] Answer: \answerYes{}{} 
    \item[] Justification: We have discuss the theoretical settings in \cref{ProofMain}

    \item {\bf Experimental result reproducibility}
    \item[] Question: Does the paper fully disclose all the information needed to reproduce the main experimental results of the paper to the extent that it affects the main claims and/or conclusions of the paper (regardless of whether the code and data are provided or not)?
    \item[] Answer: \answerYes{} 
    \item[] Justification: We discuss the experiment settings in \cref{experiment}

\item {\bf Open access to data and code}
    \item[] Question: Does the paper provide open access to the data and code, with sufficient instructions to faithfully reproduce the main experimental results, as described in supplemental material?
    \item[] Answer: \answerYes{} 
    \item[] Justification: We included the code in the submission.

\item {\bf Experimental setting/details}
    \item[] Question: Does the paper specify all the training and test details (e.g., data splits, hyperparameters, how they were chosen, type of optimizer, etc.) necessary to understand the results?
    \item[] Answer: \answerYes{} 
    \item[] Justification: The experiment details are in the code.

\item {\bf Experiment statistical significance}
    \item[] Question: Does the paper report error bars suitably and correctly defined or other appropriate information about the statistical significance of the experiments?
    \item[] Answer: \answerYes{} 
    \item[] Justification: The experiments are run on multiple seeds.

\item {\bf Experiments compute resources}
    \item[] Question: For each experiment, does the paper provide sufficient information on the computer resources (type of compute workers, memory, time of execution) needed to reproduce the experiments?
    \item[] Answer: \answerYes{} 
    \item[] Justification: We provided the related information in \cref{experiment}
    
\item {\bf Code of ethics}
    \item[] Question: Does the research conducted in the paper conform, in every respect, with the NeurIPS Code of Ethics \url{https://neurips.cc/public/EthicsGuidelines}?
    \item[] Answer: \answerYes{} 
    \item[] Justification: Code of ethics are followed.

\item {\bf Broader impacts}
    \item[] Question: Does the paper discuss both potential positive societal impacts and negative societal impacts of the work performed?
    \item[] Answer: \answerNA{} 
    \item[] Justification: There is no societal impact.
    
\item {\bf Safeguards}
    \item[] Question: Does the paper describe safeguards that have been put in place for responsible release of data or models that have a high risk for misuse (e.g., pretrained language models, image generators, or scraped datasets)?
    \item[] Answer: \answerNA{} 
    \item[] Justification: Do not have such risks.

\item {\bf Licenses for existing assets}
    \item[] Question: Are the creators or original owners of assets (e.g., code, data, models), used in the paper, properly credited and are the license and terms of use explicitly mentioned and properly respected?
    \item[] Answer: \answerNA{} 
    \item[] Justification: We do not used existing assets.

\item {\bf New assets}
    \item[] Question: Are new assets introduced in the paper well documented and is the documentation provided alongside the assets?
    \item[] Answer: \answerYes{} 
    \item[] Justification: We have included the code to verify our theorems.

\item {\bf Crowdsourcing and research with human subjects}
    \item[] Question: For crowdsourcing experiments and research with human subjects, does the paper include the full text of instructions given to participants and screenshots, if applicable, as well as details about compensation (if any)? 
    \item[] Answer: \answerNA{} 
    \item[] Justification:  The paper does not involve crowdsourcing nor research with human subjects.

\item {\bf Institutional review board (IRB) approvals or equivalent for research with human subjects}
    \item[] Question: Does the paper describe potential risks incurred by study participants, whether such risks were disclosed to the subjects, and whether Institutional Review Board (IRB) approvals (or an equivalent approval/review based on the requirements of your country or institution) were obtained?
    \item[] Answer: \answerNA{} 
    \item[] Justification:  The paper does not involve crowdsourcing nor research with human subjects.

\item {\bf Declaration of LLM usage}
    \item[] Question: Does the paper describe the usage of LLMs if it is an important, original, or non-standard component of the core methods in this research? Note that if the LLM is used only for writing, editing, or formatting purposes and does not impact the core methodology, scientific rigorousness, or originality of the research, declaration is not required.
    \item[] Answer: \answerNA{} 
    \item[] Justification: The core method does not involve LLMs.

\end{enumerate}

\end{document}